\documentclass[runningheads]{llncs}
\usepackage{graphicx}
\usepackage{textcomp}
\usepackage{xcolor}
\usepackage{algorithm}
\usepackage{algpseudocode}
\usepackage{longtable}
\usepackage{supertabular}
\usepackage{subcaption} 
\usepackage[super]{nth}

\begin{document}
\title{A Comparative Study of Sentence Embedding Models for Assessing Semantic Variation\thanks{This work was partially supported by Army Research Office Grant No. W911NF-20-1-0213.}}

\titlerunning{A Comparative Study of Sentence Embedding Models}

\author{Deven M. Mistry\inst{1}\orcidID{0000-0001-5084-8112} \and
Ali A. Minai\inst{1}\orcidID{0000-0001-9727-1701}}

\authorrunning{D.M. Mistry and A.A. Minai}

\institute{University of Cincinnati, Cincinnati, OH 45221-0030, USA
\\
\email{mistryds@mail.uc.edu, ali.minai@uc.edu}}

\maketitle             
\begin{abstract}
Analyzing the pattern of semantic variation in long real-world texts such as books or transcripts is interesting from the stylistic, cognitive, and linguistic perspectives. It is also useful for applications such as text segmentation, document summarization, and detection of semantic novelty.
The recent emergence of several vector-space methods for sentence embedding has made such analysis feasible. However, this raises the issue of how consistent and meaningful the semantic representations produced by various methods are in themselves. In this paper, we compare several recent sentence embedding methods via time-series of semantic similarity between successive sentences and matrices of pairwise sentence similarity for multiple books of literature. In contrast to previous work using target tasks and curated datasets to compare sentence embedding methods, our approach provides an evaluation of the methods “in the wild”. We find that most of the sentence embedding methods considered do infer highly correlated patterns of semantic similarity in a given document, but show interesting differences.
\end{abstract}
\section{Introduction}
The semantic structure of natural real-world texts –- especially long documents such as books –- is interesting for several reasons. Since the text is the result of a compositional cognitive process, the pattern of sequential semantic variation in it gives clues about that process. The global pattern of semantic relationships can also characterize the style, type, and content of the document (e.g., the plot of a novel). Semantic structure is also useful as the basis of semantic segmentation \cite{choi-2000-advances,Riedl2012textsegmentation,Alemi2015TextSB}, which is needed for many NLP applications.

The motivating application for the work in this paper is the identification of unusual or novel statements in texts, but the study takes a more general approach that can be useful in other ways as well. To this end, we compare eight recent sentence representation methods on several literary texts to assess how mutually consistent the semantic representations inferred by each method are over a set of long text documents. This study is not a hypothesis-driven investigation but a comparison study to assess whether and how much different representation models agree on complex, real-world texts, since they all claim to capture the actual meaning of texts. 

\section{Motivation}
The success of recently developed deep learning-based models for sentence representation \cite{hill-etal-2016-learning,cer-etal-2017-semeval,Arora:2017,conneau-kiela-2018-senteval} on systematic tests reveals their utility, but does not demonstrate whether they detect the {\it same} semantic relationships in a text, or how semantically accurate they are per se. Typically, the tests –- including those directly inferring semantic similarity between labeled sentence pairs \cite{conneau-kiela-2018-senteval,cer-etal-2017-semeval} –- use carefully curated benchmark datasets. Alternatively, performance on downstream benchmark tasks is used to evaluate the quality of sentence representations. These controlled evaluation methods are very valuable but limited by their constraints -- as is the case with most laboratory studies. The present study takes a complementary approach by looking directly at the structure of semantic variation inferred by various methods on several large real-world documents with a complex semantic structure, i.e., literary books.

Since the texts are not specially constructed or selected to fit the evaluative task (e.g., sets of labeled sentence pairs or items from different newsgroups), but are real-world documents used {\it as found}, we term this approach as evaluation ``in the wild'' (as opposed to evaluation in the lab.) While this complicates the process of evaluation, it provides a more realistic assessment of how the various computational models fare when they encounter truly natural texts.

\section{Conceptual Framework}
Foregoing the use of curated benchmarks, labeled data, and downstream tasks necessitates the adoption of a new evaluative method based on some {\it intrinsic} aspect of the results obtained. In this study, we propose and use a framework based on the following sequence of postulates:

\begin{enumerate}
    \item Every document has a specific (but latent) \emph{intrinsic meaning} and any effective semantic representation method must capture this.
    
    \item A specific intrinsic meaning implies a specific \emph{semantic structure} in a document, and all effective semantic representation methods must infer the {\it same} semantic structure for a given document
    
    \item The semantic structure of a document can be represented as the \emph{pattern of semantic similarity} between the sentences of the document.
    
    \item If two sufficiently different semantic representation methods infer mutually consistent semantic structures for a document, they are both likely to be inferring its true semantic structure.
    
    \item If two semantic representation methods infer very different semantic representations for the same document, one or both must have failed to capture its intrinsic semantic structure.

\end{enumerate}

Essentially, this proposes that, while it is difficult to determine whether a given vector representation captures the intrinsic meaning of any individual sentence, the overall semantic structure of an entire document, as represented in its sentence similarity pattern, can be used as an {\it observable surrogate representation} for its meaning, and if very different semantic representation methods infer consistent structure for a document, they must be capturing the ground truth, even though the ground truth is not explicitly known. Thus, the {\it mutual consistency} of the inferred semantic structure can be used as an implicit {\it semantic cross-validation} to evaluate a group of semantic representation methods. From a practical viewpoint, if multiple methods indicate that a particular sentence or passage in the text is dissimilar to the bulk of the document, it would provide a more reliable identification of novel statements, which is our motivating application.

\section{Methods}
\subsection{Datasets}
We use a dataset comprising the following eighteen texts:

\begin{enumerate}
    \item {\it A Christmas Carol} by Charles Dickens 
    (1,942 sentences, 29,112 word tokens).
    \item {\it Heart of Darkness} by Joseph Conrad 
    (2,430 sentences, 39,061 word tokens).
    \item {\it Metamorphosis} by Franz Kafka (translated by David Wyllie, 2002 - used under Project Gutenberg License) 
    (795 sentences, 22,373 word tokens).
    \item {\it The Prophet} by Khalil Gibran 
    (647 sentences, 12,360 word tokens).
    \item {\it A Modest Proposal} by Jonathan Swift 
    (68 sentences, 3431 word tokens)
    \item {\it A Study in the Scarlet} by Arthur Conan Doyle 
    (2,689 sentences, 43,919 word tokens)
    \item {\it Adventures of Huckleberry Finn} by Mark Twain 
    (5,789 sentences, 116,313 word tokens)
    \item {\it Dragons and Cherry Blossoms} by Mrs. Robert C. Morris 
    (1,174 sentences, 29,157 word tokens)
    \item {\it Laughter: An essay on the Meaning of the Comic} by Henri Bergson 
    (1,794 sentences, 42,947 word tokens)
    \item {\it Little Women} by Louisa May Alcott 
    (9,438 sentences, 190,752  word tokens)
    \item {\it The Picture of Dorian Gray} by Oscar Wilde 
    (6,479 sentences, 79,978 word tokens)
    \item {\it Ruth of the U.S.A} by Edwin Balmer 
    (5,093 sentences, 98,880 word tokens)
    \item {\it Siddarhtha} by Hermann Hesse 
    (1,850 sentences, 39,719 word tokens)
    \item {\it The Catspaw} by  George O. Smith 
    (1,555 sentences, 19,271 word tokens)
    \item {\it The Hound Of The Baskervilles} by Arthur Conan Doyle 
    (3,876 sentences, 59,802 word tokens)
    \item {\it The Scarlet Letter} by Nathaniel Hawthorne 
    (3,500 sentences, 84,709 word tokens)
    \item {\it The Sons Of Japheth} by Richard Wilson 
    (203 sentences, 2327 word tokens)
    \item {\it Treasure Island} by Robert Louis Stevenson 
    (3,732 sentences, 70,077 word tokens)

\end{enumerate}
The main considerations in choosing these were: a) moderate length – which makes it possible to inspect the results visually; b) diversity of type; and c)literary value, so that the texts are semantically complex and the results are of general interest; and d) Availability without violation of copyright. 
All documents were downloaded from the Project Gutenberg website\\ (https://www.gutenberg.org/).

\begin{figure}[]
    \centering
    \includegraphics[width=0.7\textwidth]{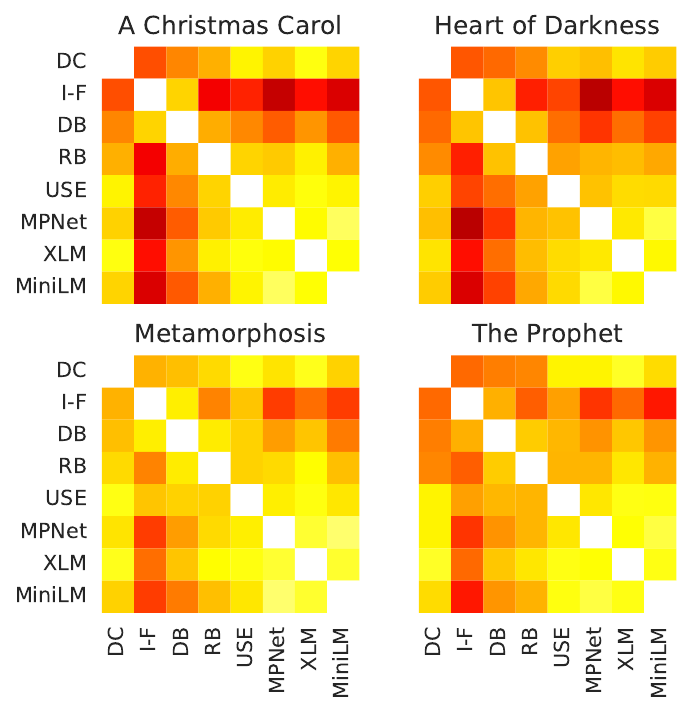}
    \caption{Correlation maps showing pairwise similarity between all methods for four books. Lighter color indicates a higher correlation.}
    \label{fig:Corr_4books}
\end{figure}

\begin{figure}[h]
    \centering
    \includegraphics[width=0.8\textwidth]{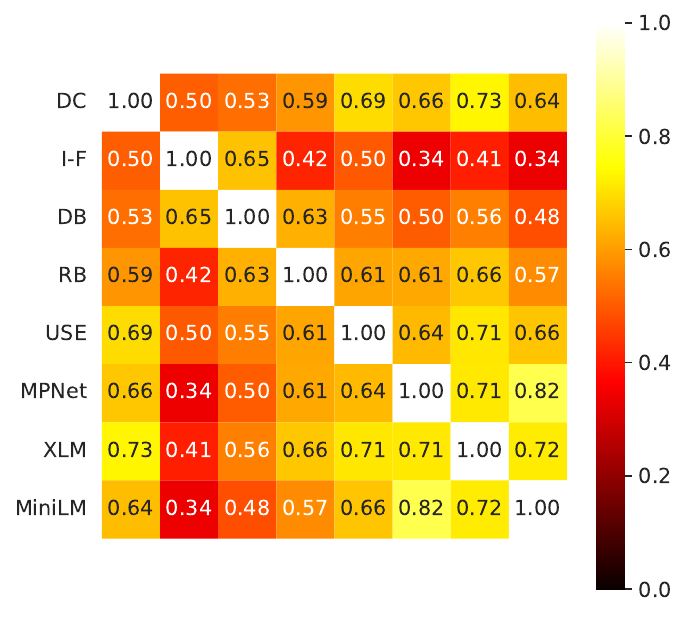}
    \caption{Mean Correlation map showing pairwise similarity between all methods for all eighteen books.}
    \label{fig:Corr_all}
\end{figure}

\begin{figure}[h]
    \captionsetup[subfigure]{aboveskip=-0.25pt, labelformat=empty}
     \centering
     \begin{subfigure}[b]{0.42\columnwidth}
         \includegraphics[width=\textwidth]{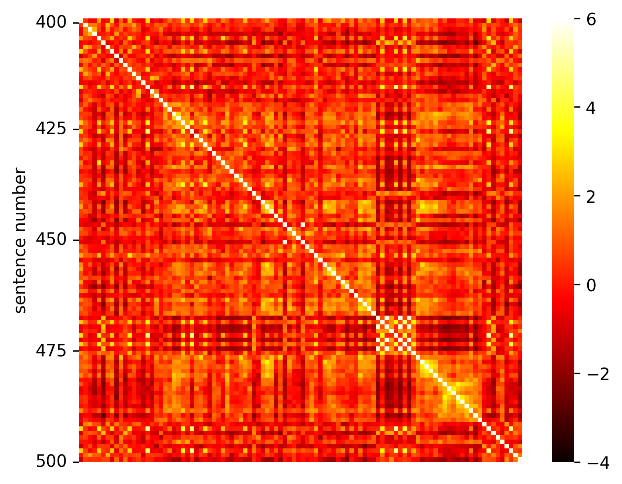}
    \label{fig:CC_DB}
     \end{subfigure}
     \begin{subfigure}[b]{0.42\columnwidth}
         \centering
         \includegraphics[width=\textwidth]{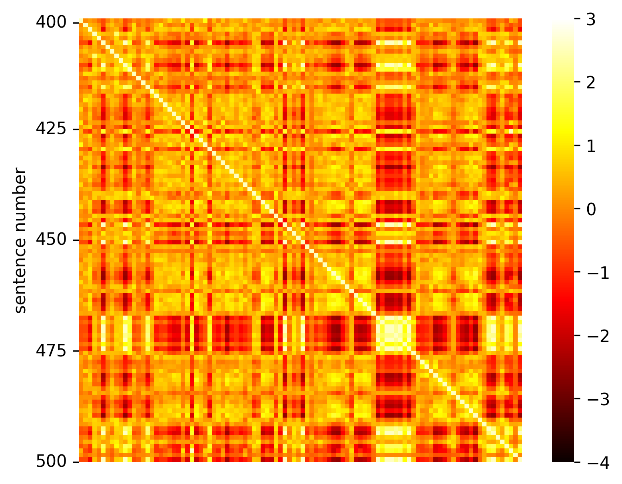}
    \label{fig:CC_IS}
     \end{subfigure}\par
     \begin{subfigure}[b]{0.42\columnwidth}
         \includegraphics[width=\textwidth]{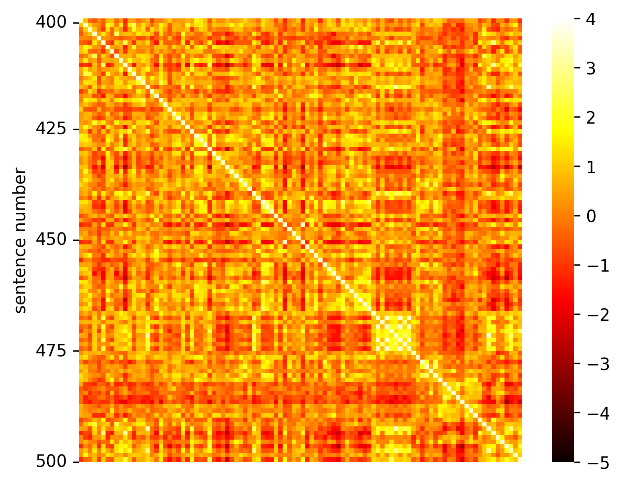}
    \label{fig:CC_DBERT}
     \end{subfigure}
     \begin{subfigure}[b]{0.42\columnwidth}
         \centering
         \includegraphics[width=\textwidth]{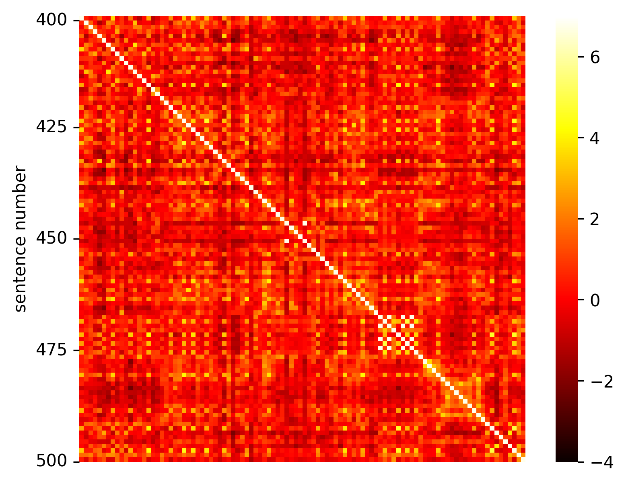}
    \label{fig:CC_MPNet}
     \end{subfigure}
\caption{Sentence similarity maps for {\it A Christmas Carol}, using DeCLUTR-Base (top left); InferSent-FasText (top right); DistilBERT (bottom left); and MPNet (bottom right).}
\label{fig:SSM}
\end{figure}

\begin{figure}[htb]
    \centering
    \includegraphics[width=\textwidth]{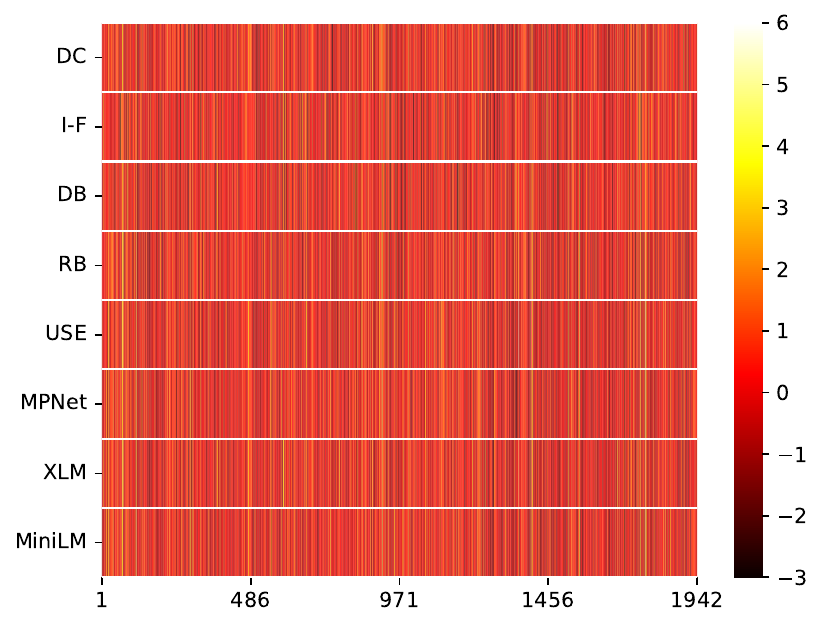}
    \caption{Time series of successive sentence similarities for {\it A Christmas Carol}.}
    \label{fig:TS_CC}
\end{figure}

\begin{figure}[]
    \captionsetup[subfigure]{aboveskip=-0.25pt, labelformat=empty}
     \centering
     \begin{subfigure}[b]{0.47\columnwidth}
         \includegraphics[width=\textwidth]{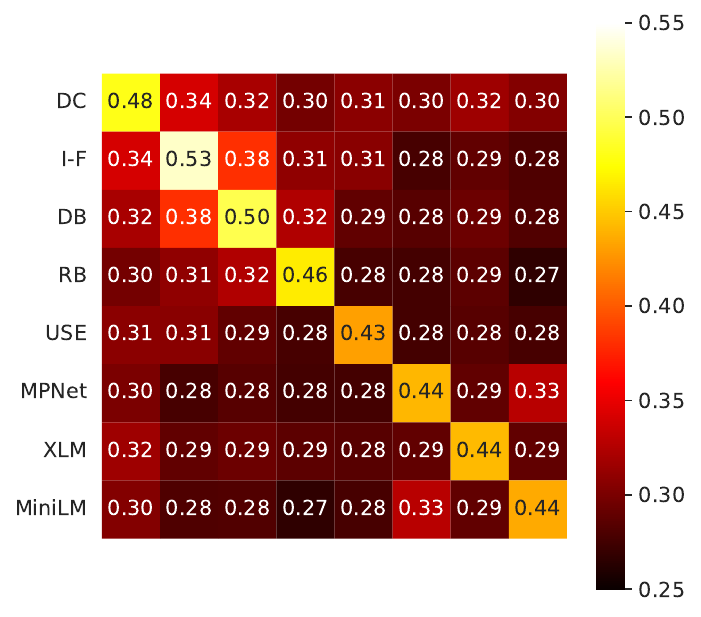}
    \label{fig:PAF}
     \end{subfigure}
     \begin{subfigure}[b]{0.47\columnwidth}
         \centering
         \includegraphics[width=\textwidth]{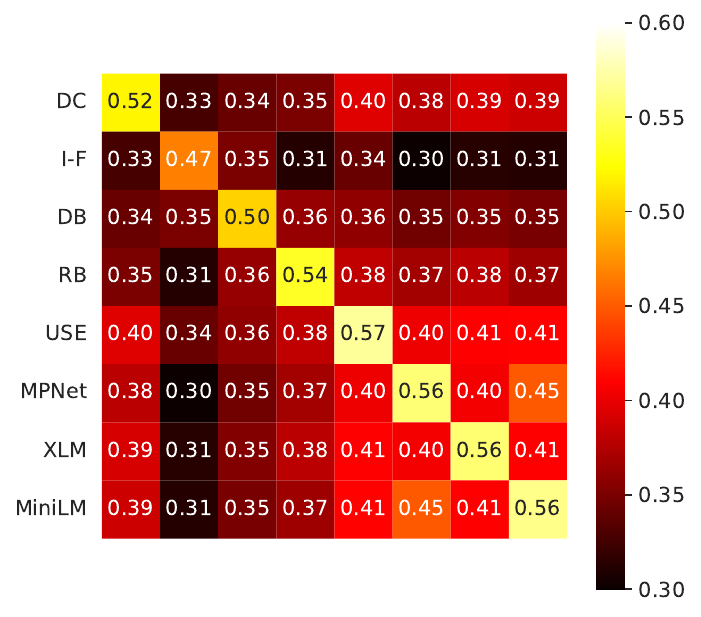}
    \label{fig:NAF}
     \end{subfigure}
\caption{Left: Positive agreement fraction (PAF) map for {\it A Christmas Carol}. Right: Negative agreement fraction (NSF) map for {\it A Christmas Carol}.}
\label{CohClust}
\end{figure}

\begin{figure}[htb]
    \centering
    \includegraphics[width=0.8\textwidth]{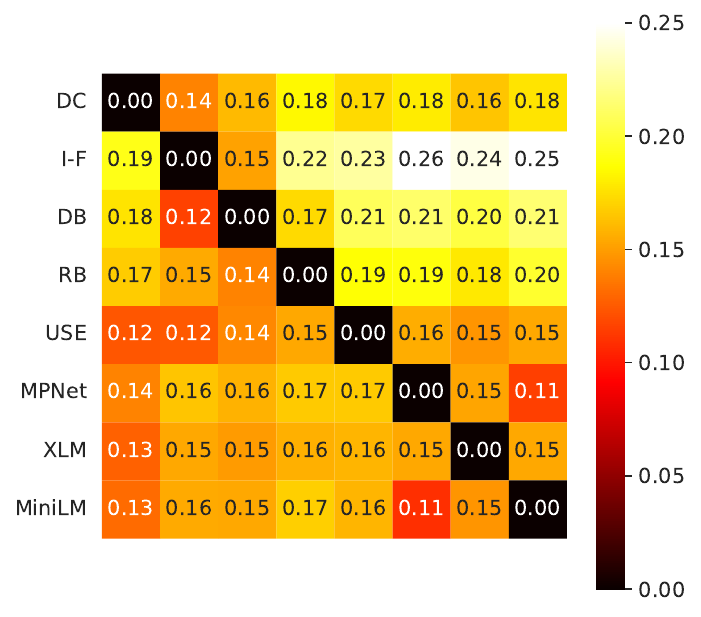}
    \caption{Directed disagreement fraction (DDAF) map for {\it A Christmas Carol}.}
    \label{fig:DDAF}
\end{figure}

\subsection{Sentence Representation Models}
It is impractical to include all the currently available sentence representation methods in our analysis, and we have tried to include a broad selection of different approaches. Specifically, the following methods are included:
\begin{enumerate}
    \item DeCLUTR Base (DC) \cite{declutr:2021}
    \item InferSent with FastText (I-F) \cite{infersent:2017}
    \item DistilBERT (DB) \cite{distilBERT:2019}
    \item RoBERTa (RB) \cite{liu2019roberta}
    \item Universal Sentence Encoder (USE) \cite{cer-etal-2018-universal}
    \item MPNet (MPNet) \cite{song2020mpnet}
    \item XLM - R (XLM) \cite{Conneau2019xlm}
    \item MiniLM (MiniLM) \cite{wang2020minilm}
    
\end{enumerate}

\noindent
The labels in parentheses are used to denote the methods in the figures.

DeCLUTR is an unsupervised learning method that explicitly uses neighboring sentences as a proxy for semantic similarity to train sentence representations. The InferSent model \cite{infersent:2017}, like DeCLUTR, is trained explicitly to represent sentence semantics, but using recurrent neural networks and supervised learning on a variety of tasks. There are versions that differ in their underlying method of representing words –- based either on  FastText word embeddings \cite{joulin-etal-2017-bag,Bojanowski2017EnrichingWV} or GloVe embeddings \cite{pennington2014glove}. The FastText version is used here. DistilBERT and RoBERTa are based on the BERT language model \cite{devlin-etal-2019-bert}. Thus, their sentence representations are tuned to the task of text-generation rather than capturing semantic similarity. The Universal Sentence Encoder (USE) model \cite{cer-etal-2018-universal} is also trained explicitly for representing sentences by training a feed-forward deep averaging network (DAN) (or a transformer) simultaneously on multiple tasks. We use the DAN version of USE, which is computationally more efficient. MiniLM \cite{wang2020minilm} proposes an effective way to compress a large transformer using deep self-distillation, where a student learns to mimic the last self-attention module of the transformer layer of the teacher. Using this approach, the trained model outperforms state-of-the-art baselines in SQuAD \cite{rajpurkar-etal-2016-squad-v1,rajpurkar-etal-2018-know-v2} and GLUE \cite{wang2018glue}. XLM-R \cite{Conneau2019xlm} is a transformer trained using masked language modeling on one hundred languages using over two terabytes of filtered CommonCrawl data. The trained model shows significant performance improvement over multilingual BERT (mBERT). MPNet \cite{song2020mpnet} adopts MLM (masked language modelling) from the original BERT model and PLM (permuted language modeling) from XLNet. The model is trained on over 160 gigabytes of data and then fine-tuned on a variety of downstream tasks to achieve better results than the existing state-of-the-art models. Given the very different architectures and training regimes of the models, it would not be surprising if they captured meaning in different ways and focused on different aspects. Demonstrating the degree and manner of this difference is a goal of this study.

\subsection{Calculating Sentence Similarity}
For each document in the corpus, the eight models listed above are used to generate embeddings for each sentence. The similarity between every pair of sentences in the document is calculated using the cosine similarity between their embeddings, thus generating an $N \times N$ {\it semantic similarity matrix} (SSM), where $N$ is the number of sentences in the document. The values in each matrix are standardized to zero-mean, unit variance values corresponding to z-scores. Thus, a negative value in cell $(i, j)$ indicates a below average similarity inferred for sentences $i$ and $j$, and a positive value indicates above average similarity within the document.

\subsection{Analysis Methods}	

We use the global pattern of semantic similarity across the entire document as captured in the SSM to evaluate and visualize the relationships between the sentence similarity patterns inferred by all the models on each given document. In addition to the SSMs, it is also interesting (and computationally simpler) to look at the time-series of similarity between successive sentences, which reflects the rhythm of meaning in the document and in the underlying generative cognitive process. 
To get a more detailed comparison, we also calculate three other metrics for each pair of models, $A$ and $B$:

\begin{enumerate}
    \item {\bf Positive Agreement Fraction (PAF):} The fraction of all sentence pairs that both model $A$ and model $B$ consider more similar than average (positive in the standardized SSMs for both models.) This matrix is symmetric, with the diagonal showing the fraction of positive sentence pairs for each model.
    
    \item {\bf Negative Agreement Fraction (NAF):} The fraction of all sentence pairs that both model $A$ and model $B$ consider less similar than average (negative in the standardized SSMs for both models.) This matrix is also symmetric, with the diagonal showing the fraction of negative sentence pairs for each model.
    
    \item {\bf Directional Disagreement Fraction (DDAF):} The fraction of all sentence pairs that model $A$ considers more similar than average (positive in the standardized SSMs for A) and model $B$ considers less similar than average (negative in the standardized SSMs for B.) This matrix is asymmetric, with the upper triangle showing the fraction of sentence pairs that are positive in $A$ and negative in $B$, and the lower triangle showing the converse.
\end{enumerate}

\section{Results and Discussion}

\subsection{Semantic Structure Comparison}

To quantify the correspondences between the SSMs generated by all the methods, we calculate the pairwise Pearson correlation coefficients between the time-series for each pair of models on each book, producing an $8 \times 8$ {\it correlation map} for each book. These are shown as heatmaps in Figure \ref{fig:Corr_4books} for four of the books. To get a more global view, these maps are averaged over all 18 documents to give the {\it mean correlation map} shown in Figure \ref{fig:Corr_all}. Several observations can be noted from these:

\begin{enumerate}
     \item Overall, a fairly similar pattern of pairwise correlation is seen in the semantic structures inferred for the four books, but the absolute level of correlation varies significantly. In general, correlations are highest for {\it Metamorphosis} and lowest for {\it The Prophet}. 
    
    \item In general, six of the methods are quite strongly correlated, with correlation coefficients well above 0.6. However, two methods -- InferSent and DistilBERT -- are less correlated with the others.
    
    \item Structures inferred by InferSent have significantly lower correlation with those inferred by the other methods except DistilBERT. The lower correlation probably reflects the fact that InferSent uses a model that is significantly different than the other methods.
    
    \item Somewhat surprisingly, DistilBERT has high correlation with both InferSent and RoBERTa. The latter is understandable, since both are BERT-based methods, but similarity with InferSent is intriguing since RoBERTa has much lower correlation with InferSent. In a sense, DistilBERT seems to bridge between InferSent and RoBERTa, agreeing with the former on some sentence pairs and agreeing with the latter on a different (though probably overlapping) set of sentence pairs.
    
    \item Interestingly, DeCLUTR has very substantial correlation with methods other than InferSent and DistilBERT even though it uses a very different approach.
    
    \item The highest correlation of any pair of methods is between MPNet and MiniLM.
    
    \item Leaving aside InferSent and DistilBERT, XLM appears to have the most correlation on average with the other four methods, which is interesting given its very different approach compared to other methods. This suggests the training on multiple languages might provide some advantage in generalization.
    
\end{enumerate}

Figure \ref{fig:SSM} shows partial SSMs obtained for {\it A Christmas Carol} using four methods. They show that these four -- and the other -- models all infer a broadly similar pattern of semantic variation in the document, though InferSent tends to assign higher similarities to sentence pairs that the other methods. In particular, the dark bands running across the maps indicate unusual or novel parts of the document, while bright patches indicate repetitive themes. While it is hard to see here, MPNet has the best fine-grained resolution in the map.

Figure  \ref{fig:TS_CC} shows the time-series of similarity between consecutive sentences generated by each model for {\it A Christmas Carol}. Visual inspection shows similarity patterns like those seen for the full SSMs, which is not surprising, since these time-series are just a plot of the first super-diagonal of each SSM. However, the degree of match between the time series is hard to appreciate visually. To look deeper, Figures \ref{CohClust} and \ref{fig:DDAF} show the PAF, NAF and DDAF values for all method pairs on {\it A Christmas Carol}. The most interesting observation from Figure \ref{CohClust} is that InferSent assigns positive (above average) similarity to more than half of the sentence pairs, DistilBERT does so for exactly half, and all the other methods assign positive similarity only to a minority of sentence pairs. This fraction is remarkably similar for USE, MPNet, XLM, and MiniLM -- all around 0.44. Another interesting observation is that in a large majority of the cases, pairs of methods agree on positive similarity for about 30\% of the sentence pairs. The clearest exception -- not surprisingly -- is InferSent. which has much higher PAF (0.38) with DistilBERT and a fairly high one (0.34) with deCLUTR. The other slight exception is a PAF of 0.33 between MPNet and MiniLM. On the NAF, InferSent has notably lower vales relative to almost all other methods, reflecting its bias towards assigning positive similarities. This is also the main reason why, in Figure \ref{fig:DDAF}, Infersent has much higher positive-to-negative disagreements with other methods than vice-versa. 

The patterns shown here for {\it A Christmas Carol}  are qualitatively similar for the other 17 books as well (not shown for lack of space).

\section{Conclusion}

This comparative study arrived at the following conclusions: 1) The semantic structure inferred for all 18 books by all the evaluated methods shows some consistency, indicating that they all partially capture the actual semantics of the document; 2) Significant differences in the semantic structure inferred by different methods indicates that each provides a distinctive take on the same document; and 3) Of the methods considered, InferSent had the lowest match with the other methods except DistilBERT, but DistilBERT also had good agreement with RoBERTa -- perhaps because both use BERT.

Based on these observations and the postulates that motivated this study, our main conclusion is that, of the 8 methods evaluated, four -- USE, MPNet, XLM, and MiniLM - provide sufficiently reliable agreement on semantic variation to be used for novelty detection. InferSent is the outlier, and its use would require much more detailed study of its biases. DeCLUTR, RoBERTa and DistilBERT fall somewhere in the middle. An interesting follow-up would to use ensembles of these methods for novelty detection.


\begin{thebibliography}{10}

\bibitem{choi-2000-advances}
Freddy Y.~Y. Choi.
\newblock Advances in domain independent linear text segmentation.
\newblock In {\em 1st Meeting of the North {A}merican Chapter of the
  Association for Computational Linguistics}, 2000.

\bibitem{Riedl2012textsegmentation}
Martin Riedl and Chris Bieman.
\newblock Text segmentation with topic models.
\newblock {\em Journal for Language Technology and Computational Linguistics},
  27(1):47--69, 2012.

\bibitem{Alemi2015TextSB}
Alexander~Amir Alemi and Paul~H. Ginsparg.
\newblock Text segmentation based on semantic word embeddings.
\newblock {\em ArXiv}, abs/1503.05543, 2015.

\bibitem{hill-etal-2016-learning}
Felix Hill, Kyunghyun Cho, and Anna Korhonen.
\newblock Learning distributed representations of sentences from unlabelled
  data.
\newblock In {\em Proceedings of the 2016 Conference of the North {A}merican
  Chapter of the Association for Computational Linguistics: Human Language
  Technologies}, pages 1367--1377, San Diego, California, June 2016.
  Association for Computational Linguistics.

\bibitem{cer-etal-2017-semeval}
Daniel Cer, Mona Diab, Eneko Agirre, I{\~n}igo Lopez-Gazpio, and Lucia Specia.
\newblock {S}em{E}val-2017 task 1: Semantic textual similarity multilingual and
  crosslingual focused evaluation.
\newblock In {\em Proceedings of the 11th International Workshop on Semantic
  Evaluation ({S}em{E}val-2017)}, pages 1--14, Vancouver, Canada, August 2017.
  Association for Computational Linguistics.

\bibitem{Arora:2017}
Sanjeev Arora, Yingyu Liang, and Tengyu Ma.
\newblock A simple but tough-to-beat baseline for sentence embeddings.
\newblock In {\em Proceedings of the International Conference on Learning
  Representations}, 2017.

\bibitem{conneau-kiela-2018-senteval}
Alexis Conneau and Douwe Kiela.
\newblock {S}ent{E}val: An evaluation toolkit for universal sentence
  representations.
\newblock In {\em Proceedings of the Eleventh International Conference on
  Language Resources and Evaluation ({LREC} 2018)}, Miyazaki, Japan, May 2018.
  European Language Resources Association (ELRA).

\bibitem{declutr:2021}
John Giorgi, Osvald Nitski, Bo~Wang, and Gary Bader.
\newblock {D}e{CLUTR}: Deep contrastive learning for unsupervised textual
  representations.
\newblock In {\em Proceedings of the 59th Annual Meeting of the Association for
  Computational Linguistics and the 11th International Joint Conference on
  Natural Language Processing (Volume 1: Long Papers)}, pages 879--895, Online,
  August 2021. Association for Computational Linguistics.

\bibitem{infersent:2017}
Alexis Conneau, Douwe Kiela, Holger Schwenk, Lo{\"\i}c Barrault, and Antoine
  Bordes.
\newblock Supervised learning of universal sentence representations from
  natural language inference data.
\newblock In {\em Proceedings of the 2017 Conference on Empirical Methods in
  Natural Language Processing}, pages 670--680, Copenhagen, Denmark, September
  2017. Association for Computational Linguistics.

\bibitem{distilBERT:2019}
Victor Sanh, Lysandre Debut, Julien Chaumond, and Thomas Wolf.
\newblock Distilbert, a distilled version of bert: smaller, faster, cheaper and
  lighter.
\newblock {\em ArXiv}, abs/1910.01108, 2019.

\bibitem{liu2019roberta}
Yinhan Liu, Myle Ott, Naman Goyal, Jingfei Du, Mandar Joshi, Danqi Chen, Omer
  Levy, Mike Lewis, Luke Zettlemoyer, and Veselin Stoyanov.
\newblock Roberta: A robustly optimized bert pretraining approach, 2019.

\bibitem{cer-etal-2018-universal}
Daniel Cer, Yinfei Yang, Sheng-yi Kong, Nan Hua, Nicole Limtiaco, Rhomni
  St.~John, Noah Constant, Mario Guajardo-Cespedes, Steve Yuan, Chris Tar,
  Brian Strope, and Ray Kurzweil.
\newblock Universal sentence encoder for {E}nglish.
\newblock In {\em Proceedings of the 2018 Conference on Empirical Methods in
  Natural Language Processing: System Demonstrations}, pages 169--174,
  Brussels, Belgium, November 2018. Association for Computational Linguistics.

\bibitem{song2020mpnet}
Kaitao Song, Xu~Tan, Tao Qin, Jianfeng Lu, and Tie-Yan Liu.
\newblock Mpnet: Masked and permuted pre-training for language understanding.
\newblock {\em arXiv preprint arXiv:2004.09297}, 2020.

\bibitem{Conneau2019xlm}
Alexis Conneau, Kartikay Khandelwal, Naman Goyal, Vishrav Chaudhary, Guillaume
  Wenzek, Francisco Guzm{\'{a}}n, Edouard Grave, Myle Ott, Luke Zettlemoyer,
  and Veselin Stoyanov.
\newblock Unsupervised cross-lingual representation learning at scale.
\newblock {\em CoRR}, abs/1911.02116, 2019.

\bibitem{wang2020minilm}
Wenhui Wang, Furu Wei, Li~Dong, Hangbo Bao, Nan Yang, and Ming Zhou.
\newblock Minilm: Deep self-attention distillation for task-agnostic
  compression of pre-trained transformers, 2020.

\bibitem{joulin-etal-2017-bag}
Armand Joulin, Edouard Grave, Piotr Bojanowski, and Tomas Mikolov.
\newblock Bag of tricks for efficient text classification.
\newblock In {\em Proceedings of the 15th Conference of the {E}uropean Chapter
  of the Association for Computational Linguistics: Volume 2, Short Papers},
  pages 427--431, Valencia, Spain, April 2017. Association for Computational
  Linguistics.

\bibitem{Bojanowski2017EnrichingWV}
Piotr Bojanowski, Edouard Grave, Armand Joulin, and Tomas Mikolov.
\newblock Enriching word vectors with subword information.
\newblock {\em Transactions of the Association for Computational Linguistics},
  5:135--146, 2017.

\bibitem{pennington2014glove}
Jeffrey Pennington, Richard Socher, and Christopher~D. Manning.
\newblock Glove: Global vectors for word representation.
\newblock In {\em Empirical Methods in Natural Language Processing (EMNLP)},
  pages 1532--1543, 2014.

\bibitem{devlin-etal-2019-bert}
Jacob Devlin, Ming-Wei Chang, Kenton Lee, and Kristina Toutanova.
\newblock {BERT}: Pre-training of deep bidirectional transformers for language
  understanding.
\newblock In {\em Proceedings of the 2019 Conference of the North {A}merican
  Chapter of the Association for Computational Linguistics: Human Language
  Technologies, Volume 1 (Long and Short Papers)}, pages 4171--4186,
  Minneapolis, Minnesota, June 2019. Association for Computational Linguistics.

\bibitem{rajpurkar-etal-2016-squad-v1}
Pranav Rajpurkar, Jian Zhang, Konstantin Lopyrev, and Percy Liang.
\newblock {SQ}u{AD}: 100,000+ questions for machine comprehension of text.
\newblock In {\em Proceedings of the 2016 Conference on Empirical Methods in
  Natural Language Processing}, pages 2383--2392, Austin, Texas, November 2016.
  Association for Computational Linguistics.

\bibitem{rajpurkar-etal-2018-know-v2}
Pranav Rajpurkar, Robin Jia, and Percy Liang.
\newblock Know what you don{'}t know: Unanswerable questions for {SQ}u{AD}.
\newblock In {\em Proceedings of the 56th Annual Meeting of the Association for
  Computational Linguistics (Volume 2: Short Papers)}, pages 784--789,
  Melbourne, Australia, July 2018. Association for Computational Linguistics.

\bibitem{wang2018glue}
Alex Wang, Amanpreet Singh, Julian Michael, Felix Hill, Omer Levy, and
  Samuel~R. Bowman.
\newblock {GLUE}: A multi-task benchmark and analysis platform for natural
  language understanding.
\newblock In {\em International Conference on Learning Representations}, 2019.

\end{thebibliography}
\end{document}